%% file: main.tex
\def\year{2021}\relax
\documentclass[letterpaper]{article} 
\pdfoutput=1
\usepackage{aaai21}  
\usepackage{times}  
\usepackage{helvet} 
\usepackage{courier}  
\usepackage[hyphens]{url}  
\usepackage{graphicx} 
\usepackage{natbib}  
\usepackage{caption} 
\usepackage{amssymb}
\usepackage{amsmath}
\usepackage{dsfont}
\usepackage{bm}
\usepackage{diagbox}
\usepackage{helvet}
\usepackage{latexsym}
\usepackage{multirow}
\usepackage[percent]{overpic}
\usepackage{subcaption}
\usepackage{tabularx}
\usepackage[table,xcdraw]{xcolor}
\usepackage[switch]{lineno}
\usepackage{booktabs}

\title{Joint Demosaicking and Denoising in the Wild:\\
The Case of Training Under Ground Truth Uncertainty}
\author{
    Jierun Chen, Song Wen, S.-H. Gary Chan\\
}
\affiliations{
    Department of Computer Science and Engineering\\
    The Hong Kong University of Science and Technology, Hong Kong, China\\
    \{jcheneh, swenaa, gchan\}@cse.ust.hk
}
\begin{document}
\maketitle

\begin{abstract}

Image demosaicking and denoising are the two key fundamental steps in digital camera pipelines, aiming to reconstruct clean color images from noisy luminance readings. In this paper, we propose and study Wild-JDD, a novel learning framework for joint demosaicking and denoising in the wild. In contrast to previous works which generally assume the ground truth of training data is a perfect reflection of the reality, we consider here the more common imperfect case of ground truth uncertainty in the wild. We first illustrate its manifestation as various kinds of artifacts including zipper effect, color moire and residual noise. Then we formulate a two-stage data degradation process to capture such ground truth uncertainty, where a conjugate prior distribution is imposed upon a base distribution. After that, we derive an evidence lower bound (ELBO) loss to train a neural network that approximates the parameters of the conjugate prior distribution conditioned on the degraded input. Finally, to further enhance the performance for out-of-distribution input, we design a simple but effective fine-tuning strategy by taking the input as a weakly informative prior. Taking into account ground truth uncertainty, Wild-JDD enjoys good interpretability during optimization. Extensive experiments validate that it outperforms state-of-the-art schemes on joint demosaicking and denoising tasks on both synthetic and realistic raw datasets. 
\end{abstract}

\section{Introduction}

Modern digital cameras use a single sensor overlaid with a color filter array (CFA) to capture an image. This means that only one color channel's value is recorded for each pixel location. Let $N$ be the number of pixels in an image, the raw data acquisition process can be simply modeled as

\begin{equation}
    \bm{x} = \bm{Az} + \bm{n},
\end{equation}
where $\bm{x} \in \mathbb{R}^N$ is a noisy raw data vector of luminance readings, $\bm{A} \in \mathbb{R}^{N\times3N}$ is a mosaicking operation, $\bm{z} \in \mathbb{R}^{3N}$ is an unknown clean image with three color channels, and $\bm{n} \in \mathbb{R}^{N}$ is a noise vector.\par

\begin{figure}
     \centering
     \begin{subfigure}[b]{0.326\linewidth}
         \centering
        \includegraphics[width=\linewidth]{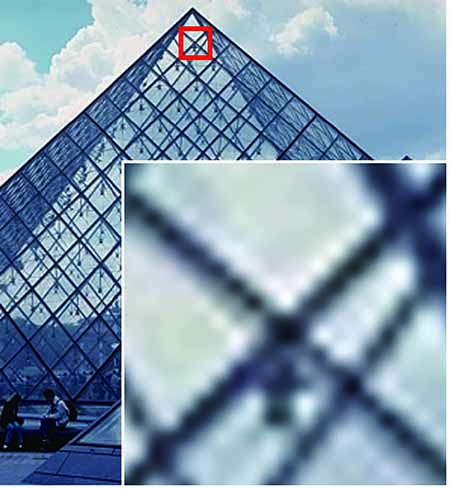}
         \caption{Zipper effect.}
     \end{subfigure}
     \hfill
     \begin{subfigure}[b]{0.327\linewidth}
         \centering
         \includegraphics[width=\linewidth]{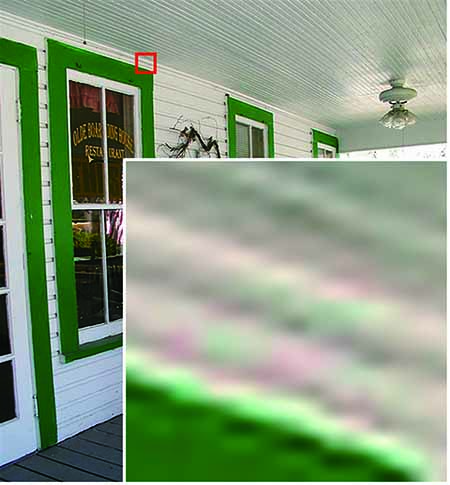}
         \caption{Color moire.}
     \end{subfigure}
     \hfill
     \begin{subfigure}[b]{0.327\linewidth}
         \centering
         \includegraphics[width=\linewidth]{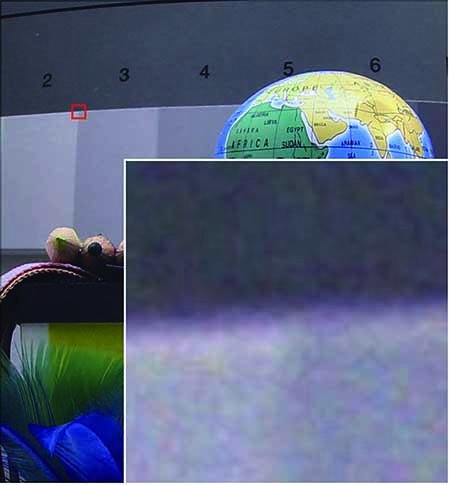}
         \caption{Residual noise.}
     \end{subfigure}
    \caption{
    Imperfect ground truth examples (\textit{electronic zoom-in recommended}): (a) A ground truth image from CBSD dataset \cite{arbelaez2011contour} suffering from zipper effect, an artificial jagged pattern around edges; (b) Color moire in an image from ImageNet dataset \cite{russakovsky2015imagenet}. Such artifact appears as false coloring due to interpolation error; (c) Noticeable residual noise in the collected ``clean'' image from Renoir dataset \cite{anaya2018renoir}.
    }
    \label{fig.artifacts}
\end{figure}

Before the final ``cooked'' image is ready for the users, the raw data undergoes a series of processing steps, known as the image processing pipeline. Among those, demosaicking and denoising (DM\&DN) are two of the very early and most crucial steps. Demosaicking aims to undo the mosaicking operation $\bm{A}$  by interpolating the missing two-thirds of each pixel's color channels, while denoising removes the inevitable noise $\bm{n}$ from the measurement $\bm{x}$. Due to their modular property, substantial traditional literature takes them as independent tasks and executes them in a sequential manner. This yields potentially suboptimal performance,  and inspires several works  on jointly addressing the DM\&DN tasks \cite{liu2020joint,kokkinos2019iterative,tan2017joint}.\par

Among the joint DM\&DN works, data-driven approaches \cite{liu2020joint,tan2018deep,kokkinos2018deep} have been shown more effective than applying handcrafted priors and filters. These approaches usually require a collection of paired data, which are the mosaicked noisy images $\bm{x}$ and the demosaicked clean ``ground truth'' counterparts $\bm{y}$. However, it is often costly and tedious to collect a large amount of high quality real-life data. Furthermore, the collected  $\bm{y}$ is not perfect without artifacts or noise. We illustrate this in Figure \ref{fig.artifacts}. For demosaicking, many approaches \cite{syu2018learning,tan2017color} take the output from a camera pipeline as $\bm{y}$, possibly introducing artifacts like zipper effect or color moire in regions with rich textures and sharp edges. For denoising, the ``clean'' images are often collected by either  setting a low-ISO \cite{plotz2017benchmarking,anaya2018renoir} or averaging a set of repeated shots of the same scene \cite{abdelhamed2018high}, which still contain noticeable noise. Moreover, such denoising data collecting process usually assumes the captured objects to be perfectly still, or requires a precise spatial alignment and intensity calibration among a burst of images. Potential failure cases would introduce additional error into the collected dataset. Therefore, all these in-the-wild issues means that the ``ground truth'' $\bm{y}$ deviates from the needed authentic  $\bm{z}$, limiting the performance of DM\&DN model.\par

To account for the fact that the collected ground truth $\bm{y}$ is not a perfect reflection of $\bm{z}$,  we propose Wild-JDD, a novel joint demosaiking and denoising learning framework to enable training under ground truth uncertainty. In Wild-JDD, we first formulate a two-stage data degradation process, where a conjugate prior distribution is imposed upon a base Gaussian distribution. Then, we derive an ELBO loss from a variational perspective. In this way, the optimization process is aware of the target uncertainty and prevents the trained neural network from over-fitting to those randomness errors. Beyond that, when the testing image falls outside of the training range, we further enhance the performance by regarding the input as a weakly informative prior. \par

Our main contributions are summarized as follows:
\begin{itemize}
\item We identify in existing DM\&DN datasets the ground truth uncertainty issues, manifesting themselves as various artifacts in the wild, such as zipper effect, color moire and residual noise.
\item We introduce a novel learning framework for joint demosaicking and denoising in the wild (Wild-JDD), where a two-stage data degradation and an ELBO loss are formulated for optimization. We also propose a simple but effective fine-tuning strategy for out-of-distribution input.
\item Instead of simply generating a demosaicked clean image, networks instantiated from our framework are capable of estimating all the parameters involved in data degradation and reconstruction, which provides better interpretability of the optimization process. 
\item We conduct extensive experiments on both synthetic and realistic datasets. Quantitative and qualitative comparisons show that Wild-JDD substantially outperforms state-of-the-art works.
\end{itemize}

\section{Related Work}
In this section, we review the most relevant DM\&DN works from sequential processing to joint optimization, and from supervised learning to self-supervised fine-tuning. \par

Traditionally, demosaicking and denoising are performed sequentially in arbitrary order~\cite{zhang2011color,akiyama2015pseudo,zhang2014joint}.
However, demosaicking first would correlate the noise spatially, break the commonly used independent identically distributed (i.i.d.) assumption imposed on the noise modeling, and increase the difficulty of the following denoising step \cite{nam2016holistic,zhang2009pca}. Another issue arises if denoising is performed first, ending up with an over-smoothed result \cite{jin2020review}. To address the above problems, recent studies jointly consider DM\&DN for better performance. \citeauthor{khashabi2014joint} proposed one of the very first joint approaches through a learned non-parametric random field, and published an MSR dataset for evaluation. \citeauthor{heide2014flexisp} embedded a non-local natural prior into a global primal-dual optimization. \citeauthor{klatzer2016learning} formulated a sequential energy minimization framework. However, these heuristics-based methods were outperformed by the deep-learning-based approaches. \citeauthor{gharbi2016deep} trained a neural network on millions of images to achieve better results and shorter running time. After that, new approaches were developed to extend the CNN's capability in the field: first with more effective network blocks \cite{huang2018lightweight,tan2018deep}, or a cascade of refinement frameworks \cite{kokkinos2019iterative,kokkinos2018deep}, then by relying on a variational deep image prior \cite{park2020joint}, and exploiting density map guidance to adaptively recover regions with different frequencies \cite{liu2020joint}. These learning-based methods have achieved state-of-art performances, but the assumption that the clean color image is perfect remains in doubt. \par

Data-driven approaches normally perform well if the testing image shares a similar distribution with the training data \cite{mohseni2020self,ehret2019joint}. However, in practice, the noise type remains diversified and may fall outside the training range. For this reason, \citeauthor{lehtinen2018noise2noise} introduced a pioneering ``noise2noise'' training strategy using pairs of noisy images. A similar ``mosaic2mosaic'' framework improved a joint DM\&DN network by fine-tuning bursts of raw images \cite{ehret2019joint}. Nonetheless, performance of these methods can be limited by inadequate shots of the same scene. \citeauthor{batson2019noise2self,krull2019noise2void} tackled the problem by using only one realization for each image at the price of a performance drop. These insightful methods enable adaptive fine-tuning, but the quality or uncertainty level of the pseudo ground truth has not been taken into account. \par

Our method differs with previous works in that we do not ideally assume the collected ``ground truth'' as perfect data. Instead, we acknowledge the presence of various artifacts and consider the case of training a joint DM\&DN network under the ground truth uncertainty.

\section{Wild-JDD Methodology}
We start with a dataset $\mathcal{D} = \left\{\bm{x}^{(i)}, \bm{y}^{(i)}\right\}_{i=1}^M$, consisting of $M$ pairs of images. Our goal is to learn a function that maps $\bm{x}^{(i)}$ to its corresponding authentic ground truth $\bm{z}^{(i)}$ (note that $\bm{y}^{(i)}$ is an approximation of $\bm{z}^{(i)}$). This goal is approached by two steps: (1) we formulate a two-stage data degradation to link up all the parameters involved; (2) we derive an expression of data likelihood for optimization. Moreover, a fine-tuning strategy is applied to deal with out-of-distribution inputs. To keep dimensional consistency across different variables, in the following, without otherwise stated, $\bm{x}^{(i)}$ is first bilinearly interpolated to a color image $\bm{\tilde{x}}^{(i)} \in \mathbb{R}^{3N}$, e.g. interpolating a missing green channel by averaging its four green neighbors. The superscript $(i)$ is ignored in following subsections for simplicity.\par

\begin{figure}
\centering
\includegraphics[width=\linewidth]{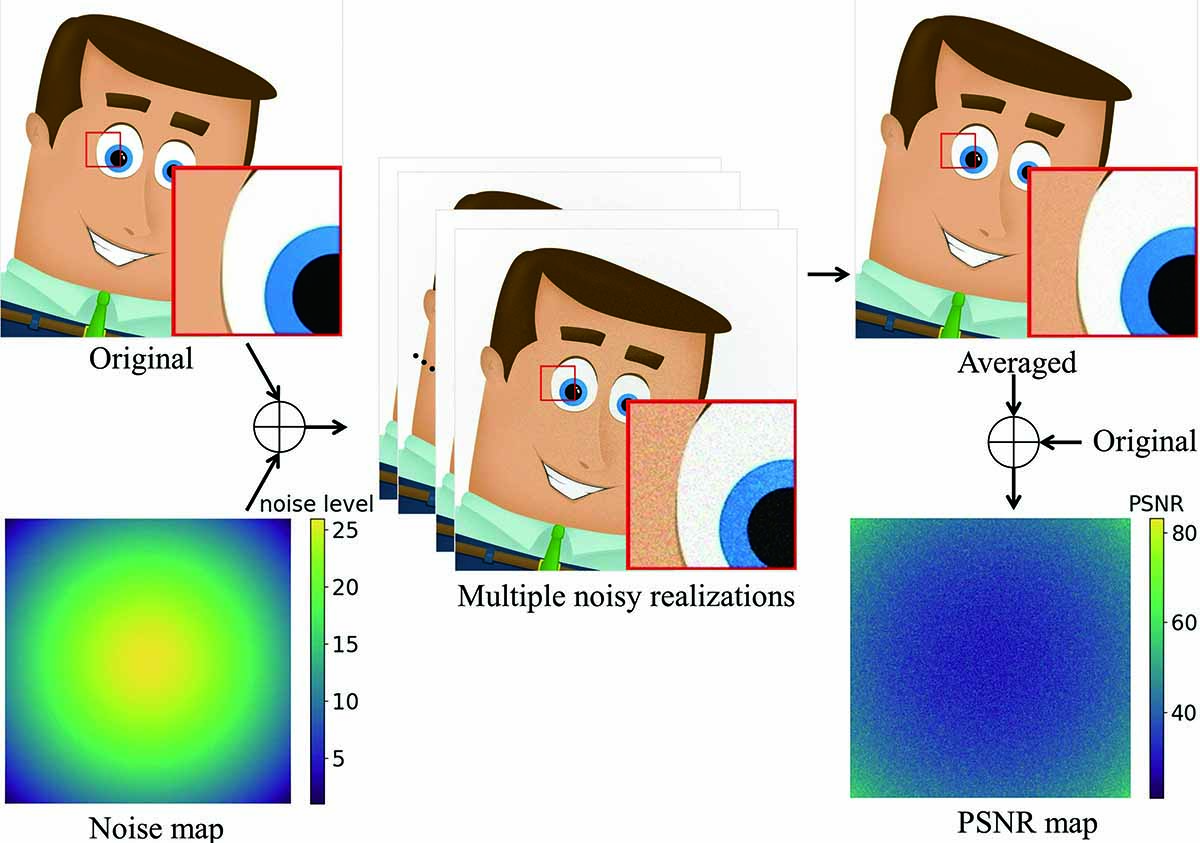}
\caption{Visualization of the relation between the quality of collected clean image and the noise level. Firstly multiple (ten here) noisy realizations are generated given the original cartoon image and its spatial noise map. Then the relatively clean image is collected by averaging across those noisy realizations. The PSNR map between the averaged image and the original version indicates that those regions with higher noise levels correspond to lower PSNR values and higher uncertainties of the collected clean pixels.}
\label{fig.m_v_relation}
\end{figure}

\subsection{Two-stage Data Degradation}
Conventionally, the pixel/channel-wise expression for data degradation is modelled with an additive Gaussian noise~\cite{zhou2019adaptation, jia2019focnet} as
\begin{equation}
    \tilde{x}_j|y_j, \sigma_j^{2} \sim \mathcal{N}(y_j, \sigma_j^{2}),
    \label{eq:normal_previous}
\end{equation}
where $j =1, 2,\cdots, 3N$ specifies a dimension within an image. However, $y_j$ is just a point estimator of the unknown authentic ground truth $z_j$, and training with $y_j$ only achieves a suboptimal performance as variance of this estimator is not considered. As such, we adopt a brand-new degradation model as 
\begin{equation}
    \tilde{x}_j|z_j, \sigma_j^{2} \sim \mathcal{N}(z_j, \sigma_j^{2}),
    \label{eq:normal}
\end{equation}
and seek to first parameterize the unknown authentic ground truth $z_j$. Suppose $z_j$ follows a normal distribution with mean $y_j$ and variance $\sigma_j^2/\lambda$:

\begin{equation}
    z_j|y_j, \sigma_j^{2}, \lambda \sim \mathcal{N}(y_j, \sigma_j^{2}/\lambda),
    \label{eq:normal z_j}
\end{equation}
where $\sigma_j^2$ has an inverse gamma distribution parameterized by $\alpha, \beta_j$ :
\begin{equation}
    \sigma_j^2|\alpha, \beta_j \sim \Gamma^{-1}(\alpha, \beta_j).
    \label{eq:IG}
\end{equation}
Then $\left(z_j, \sigma_j^2\right)$ can be jointly denoted as a normal-inverse-gamma distribution:

\begin{equation}
\left(z_j, \sigma_j^{2}\right)|y_j, \lambda, \alpha, \beta_j \sim \mathrm{N}\text{-}\Gamma^{-1}(y_j, \lambda, \alpha, \beta_j),
\label{eq:NIG}
\end{equation}
which serves as a conjugate prior distribution over the base distribution in Equation (\ref{eq:normal}). The collected ground truth $y_j$ is an estimate of $z_j$, while parameter $\lambda$ reflects the quality or uncertainty level of this estimation. For $\alpha$ and $\beta_j$, they can be interpreted in a way that the variance $\sigma_j^2$ is estimated from $2\alpha$ observations with a sum of sample squared deviations $2\beta_j$. Similar to estimating the noise level by applying a Gaussian filter to the variance map \cite{yue2019variational}, we parameterize $\alpha$ and $\beta_j$ as
\begin{equation}
    \begin{split}
        \alpha &= \frac{w^2}{2}, \\
        \beta_j &= \frac{w^2}{2}
                  \mathcal{B}(\{(x_{j+t}-y_{j+t})^2\}_{t=-\lfloor w^2/2 \rfloor}^{\lfloor w^2/2 \rfloor}),
    \end{split}
\end{equation}
where $\mathcal{B}$ denotes a bilateral filtering operation on a variance map patch centered at pixel $j$ with odd window size $w$. Note that $\lambda$ and $\alpha$ here are invariant to the dimension $j$. According to Equation (\ref{eq:normal z_j}), the variance of $z_j$ or the uncertainty of $y_j$ is proportional to the noise level $\sigma_j^2$. This can be reasonable, e.g., when collecting the clean ground truth by averaging multiple shots of the same scene, the noisier region should have lower quality and higher uncertainty (see Figure \ref{fig.m_v_relation}). \par
\begin{figure}
\centering
\includegraphics[width=\linewidth]{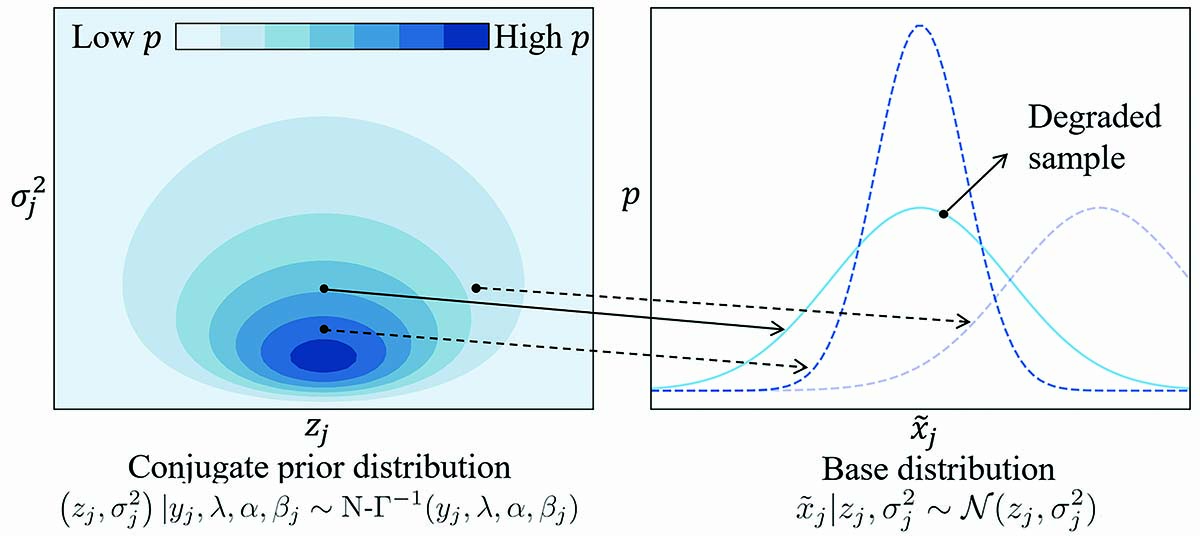}
\caption{Two-stage data degradation corresponds to two sampling processes: first sampling from the conjugate prior distribution to obtain parameters of the base distribution; then sampling from the base distribution to obtain a degraded sample.}
\label{fig.data_degradation}
\end{figure}

Therefore, based on equations (\ref{eq:normal}) and (\ref{eq:NIG}), the data degradation process comprises two stages (see Figure \ref{fig.data_degradation}): (1) $z_j$ and $\sigma_j^2$ take values by sampling from the prior distribution $p(z_j,\sigma_j^2|y_j, \lambda, \alpha, \beta_j)$; (2) then a degraded sample $\tilde{x}_j$ is generated from the conditional distribution $p(\tilde{x}_j|z_j,\sigma_j^2)$.

\begin{figure*}
\centering
\includegraphics[width=0.9\textwidth]{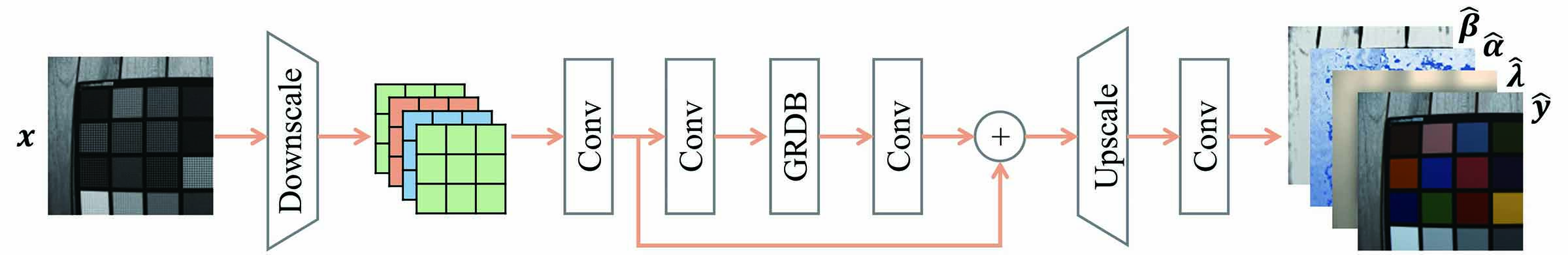}
\caption{The overall architecture. The input was first downscaled into four color maps. Then a series of convolutional layers are applied to remove the noise and to interpolate missing color values. The output is obtained by applying an upscaling operation followed by an additional convolutional layer. This network learns a posterior distribution $q_{\bm{w}}(\bm{z},\bm{\sigma}^2|\bm{x}) = \mathrm{N}\text{-}\Gamma^{-1}(\bm{\hat{y}}, \bm{ \hat{\lambda}}, \bm{ \hat{\alpha}}, \bm{ \hat{\beta}})$.}
\label{fig.network}
\end{figure*}

\subsection{Maximizing ELBO}
For the purpose of predicting latent variables $\bm{z},\bm{\sigma}^2$ given observed data point $\bm{\tilde{x}}$, or approximating the intractable posterior $p(\bm{z},\bm{\sigma}^2|\bm{\tilde{x}})$, an encoder $q_{\bm{w}}(\bm{z},\bm{\sigma}^2|\bm{\tilde{x}})$ is introduced with learnable weights $\bm{w}$, as in the work \cite{kingma2013auto}. Notably, due to our use of a conjugate prior in Equation (\ref{eq:NIG}), this encoder $q_{\bm{w}}(\bm{z},\bm{\sigma}^2|\bm{\tilde{x}})$ is in the same probability distribution family as the prior, i.e.,
\begin{equation}
    q_{\bm{w}}(\bm{z},\bm{\sigma}^2|\bm{\tilde{x}}) = \prod_{i=1}^{3N}
            \mathrm{N}\text{-}\Gamma^{-1}(\hat{y}_j, \hat{\lambda}_j, \hat{\alpha}_j, \hat{\beta}_j),
\end{equation}
where $\{\hat{y}_j, \hat{\lambda}_j, \hat{\alpha}_j, \hat{\beta}_j\}_{j=1}^{3N}$ is the output of a neural network. Note that here $\hat{\lambda}_j$ is dimension-wise output learning the same target $\lambda$ for different $j$. The same is applicable to $\hat{\alpha}_j$ and $\alpha$.\par

We next describe how we actually train such a neural network using
a maximum likelihood estimation approach. As described in the work \cite{kingma2013auto}, the marginal log-likelihood can be calculated as
\begin{equation}
    \log p(\bm{\tilde{x}}) = D_{KL}(q_{\bm{w}}(\bm{z},\bm{\sigma}^2|\bm{\tilde{x}})||p(\bm{z},\bm{\sigma}^2|\bm{\tilde{x}}))
    + \mathcal{L}(\bm{w};\bm{\tilde{x}}),
\end{equation}
where the KL divergence term is a non-negative value. Therefore, maximizing the marginal log-likelihood $\log p(\bm{\tilde{x}})$ is converted to maximizing the second term, called the evidence lower bound (ELBO), and decomposed as
\begin{equation}
     \begin{aligned}
    \mathcal{L}(\bm{w};\bm{\tilde{x}}) =
    & -D_{KL}(q_{\bm{w}}(\bm{z},\bm{\sigma}^2|\bm{\tilde{x}})||p(\bm{z},\bm{\sigma}^2))  \\
    & + \mathbb{E}_{q_{\bm{w}}(\bm{z},\bm{\sigma}^2|\bm{\tilde{x}})}\left[\log p(\bm{\tilde{x}} | \bm{z},\bm{\sigma}^2)\right].
    \end{aligned}
    \label{ELBO}
\end{equation}
This ELBO loss is maximized when: (1) the divergence term encourages the distribution returned by the encoder network close to the prior; (2) the expectation term guides the network predicting parameters with a high likelihood after seeing the corrupted image. In the work \cite{kingma2013auto}, the sampling process requires a reparameterization trick for gradient back-propagation. However, such kinds of tricks are unnecessary here because a closed form expression for $\mathcal{L}(\bm{w};\bm{\tilde{x}})$ can be derived analytically as follows:

\begin{align}
\begin{split}\label{eq.Dkl}
    &\quad D_{KL}(q_{\bm{w}}(\bm{z},\bm{\sigma}^2|\bm{\tilde{x}})||p(\bm{z},\bm{\sigma}^2)) \\
    &= \sum_{j=1}^{3N}\left\{
    \frac{\lambda \hat{\alpha}_j}{2\hat{\beta}_j}(y_j-\hat{y}_j)^2+\frac{\lambda}{2\hat{\lambda}_j}-\frac{1}{2}\log\frac{\lambda}{\hat{\lambda}_j}+\alpha_j\log{\frac{\hat{\beta}_j}{\beta_j}}\right.\\
    &\quad  \left.-\frac{1}{2}+\log{\frac{\Gamma(\alpha_j)}{\Gamma(\hat{\alpha}_j)}}+(\hat{\alpha}_j-\alpha_j)\psi(\hat{\alpha}_j)-(\hat{\beta}_j-\beta_j)\frac{\hat{\alpha}_j}{\hat{\beta}_j} \right\},
\end{split}\\
\begin{split}\label{eq.Expectation}
    &\quad \mathbb{E}_{q(\bm{z},\bm{\sigma}^2|\bm{\tilde{x}})}\left[\log p(\bm{\tilde{x}} | \bm{z},\bm{\sigma}^2)\right]=\sum_{j=1}^{3N}\left\{
    -\frac{\log{2\pi}}{2} \right.\\
    &\qquad -\frac{\log{\hat\beta_j}-\psi(\hat\alpha_j)}{2} 
    \left.
    -\frac{\hat\beta_j}{2\hat{\lambda}_j^2(\hat\alpha_j-1)}
    -\frac{\hat\alpha_j(\tilde{x}_j-\hat{y}_j)^2}{2\hat\beta_j}
    \right\},
\end{split}
\end{align}

\noindent
where $\Gamma(\cdot), \psi(\cdot)$ denote the Gamma and Di-gamma function respectively (detailed derivations are provided in the supplementary materials). Looking deeper into the term $\frac{\lambda \hat{\alpha}_j}{2\hat{\beta}_j}(y_j-\hat{y}_j)^2$ in Equation (\ref{eq.Dkl}), we can notice that if the parameter $\lambda_j$ is set to be large enough, the ELBO loss would degenerate to a mean squared error (MSE). When using an MSE loss, too much attention would be put into the restoration term $(y_j-\hat{y}_j)^2$, leaving the existence of ground truth uncertainty omitted and making the model biased to the training data. Therefore, from a variational point of view, Wild-JDD provides a sound interpretability of the reason why the restoration term and the rest regularization terms should coexist in the training process. \par

After formulating the ELBO loss for each single image, the overall optimization objective is obtained by computing across the entire dataset:
\begin{equation}
    \min_{\bm{w}} \sum_{i=1}^{M} -\log p(\bm{\tilde{x}}^{(i)}); \log p(\bm{\tilde{x}}^{(i)}) \approx \mathcal{L}(\bm{w};\bm{\tilde{x}}^{(i)}).
\end{equation}
\par
At test time, the desired demosaicked clean image is obtained by taking the expectation of $\bm{z}$, i.e. $\mathbb{E}[\bm{z}] = \bm{\hat{y}}$, while the noise map is parameterized as $\mathbb{E}[\bm{\sigma^2}]=\frac{\bm{\hat{\beta}}}{\bm{\hat{\alpha}}-1}$, according to the definition of $\mathrm{N}\text{-}\Gamma^{-1}$ distribution.

\begin{table*}[t]
\centering
\begin{tabularx}{\textwidth}{@{\extracolsep{\fill}}lccccccccccc}
\hline
 \multirow{2}{*}{Method}& \multirow{2}{*}{$\sigma$} & \multicolumn{2}{c}{\begin{tabular}[c]{@{}c@{}}Kodak\\ (24 images)\end{tabular}} & \multicolumn{2}{c}{\begin{tabular}[c]{@{}c@{}}McMaster\\ (18 images)\end{tabular}} & \multicolumn{2}{c}{\begin{tabular}[c]{@{}c@{}}WED-CDM\\ (100 images)\end{tabular}} & \multicolumn{2}{c}{\begin{tabular}[c]{@{}c@{}}MIT moire\\ (1000 images)\end{tabular}} & \multicolumn{2}{c}{\begin{tabular}[c]{@{}c@{}}Urban100\\ (100 images)\end{tabular}} \\
 &  & PSNR & SSIM & PSNR & SSIM & PSNR & SSIM & PSNR & SSIM & PSNR & SSIM \\ 
\hline
FlexISP &  & 31.31 & 0.8694 & 31.17 & 0.8627 & 31.08 & 0.8754 & 29.06 & 0.8206 & 30.37 & 0.8832 \\
SEM &  & 34.59 & 0.9269 & 32.36 & 0.8869 & 32.85 & 0.9234 & 27.46 & 0.8292 & 27.19 & 0.7813 \\
ADMM &  & 31.60 & 0.8787 & 32.63 & 0.8966 & 31.79 & 0.9003 & 28.58 & 0.7923 & 28.57 & 0.8578 \\
DeepJoint & 5 & 36.11 & 0.9455 & 35.47 & 0.9378 & 35.09 & 0.9485 & 31.82 & 0.9015 & 34.04 & 0.9510 \\
Kokkinos &  & 36.22 & 0.9426 & 34.74 & 0.9252 & 35.12 & 0.9410 & 31.94 & 0.8882 & 34.07 & 0.9358 \\
SGNet &  & - & - & - & - & - & - & 32.15 & \textbf{0.9043} & 34.54 & 0.9533 \\
Wild-JDD & \multicolumn{1}{l}{} & {\textit{36.88}}  & {\textit{0.9520}} &  {\textit{35.85}} & {\textit{0.9425}} & {\textit{35.92 }}& {\textit{0.9543}} & {\textit{32.29}} & 0.8987 & {\textit{34.70}} & {\textit{0.9534 }}\\
Wild-JDD* & {\color[HTML]{333333} } & \textbf{36.97}  & \textbf{0.9526}  & \textbf{35.94} & \textbf{0.9435} & \textbf{36.01} & \textbf{0.9551} & \textbf{32.39} & {\textit{0.8999}} & \textbf{34.83} & \textbf{0.9540} \\ 
\hline
FlexISP &  & 28.64 & 0.7583 & 28.51 & 0.7534 & 28.24 & 0.7691 & 26.61 & 0.7491 & 27.51 & 0.8196 \\
SEM &  & 29.78 & 0.7681 & 28.68 & 0.7306 & 28.90 & 0.7563 & 25.45 & 0.7531 & 25.36 & 0.7094 \\
ADMM &  & 31.04 & 0.8595 & 31.72 & 0.8699 & 30.90 & 0.8758 & 28.26 & 0.7720 & 27.48 & 0.8388 \\
DeepJoint & 10 & 33.10 & 0.9018 & 33.18 & 0.9047 & 32.69 & 0.9156 & 29.75 & 0.8561 & 31.60 & 0.9152 \\
Kokkinos &  & 33.32 & 0.9022 & 32.75 & 0.8956 & 32.76 & 0.9066 & 30.01 & 0.8123 & 31.73 & 0.8912 \\
SGNet &  & - & - & - & - & - & - & 30.09 & 0.8619 & 32.14 & 0.9229 \\
Wild-JDD & \multicolumn{1}{l}{} & {\textit{33.81}} & {\textit{0.9127}} & {\textit{33.53}} & {\textit{0.9123}} & {\textit{33.44}} & {\textit{0.9244}} & {\textit{30.30}} & {\textit{0.8645}} & {\textit{32.42}} & {\textit{0.9288}} \\
Wild-JDD* &  & \textbf{33.88} & \textbf{0.9136} & \textbf{33.61} & \textbf{0.9137} & \textbf{33.51} & \textbf{0.9255} & \textbf{30.37} & \textbf{0.8657} & \textbf{32.54} & \textbf{0.9299} \\ 
\hline
FlexISP &  & 26.67 & 0.6541 & 26.55 & 0.6572 & 26.24 & 0.6694 & 24.91 & 0.6851 & 25.55 & 0.7642 \\
SEM &  & 25.79 & 0.5954 & 25.45 & 0.5800 & 25.46 & 0.5799 & 23.23 & 0.6527 & 23.25 & 0.6156 \\
ADMM &  & 30.16 & 0.8384 & 30.50 & 0.8412 & 29.85 & 0.8497 & 27.58 & 0.7497 & 28.37 & 0.8440 \\
DeepJoint & 15 & 31.25 & 0.8603 & 31.49 & 0.8707 & 30.99 & 0.8823 & 28.22 & 0.8088 & 29.73 & 0.8802 \\
Kokkinos &  & 31.28 & 0.8674 & 30.98 & 0.8605 & 30.94 & 0.8710 & 28.28 & 0.7693 & 29.87 & 0.8451 \\
SGNet &  & - & - & - & - & - & - & 28.60 & 0.8188 & 30.37 & 0.8923 \\
Wild-JDD & \multicolumn{1}{l}{} & {\textit{31.92}} & {\textit{0.8765}} & {\textit{31.90}} & {\textit{0.8846}} & {\textit{31.75}} & {\textit{0.8965}} & {\textit{28.89}} & {\textit{0.8310}} & {\textit{30.79}} & {\textit{0.9055}} \\
Wild-JDD* &  &\textbf{31.99} & \textbf{0.8777} & \textbf{31.97} & \textbf{0.8863} & \textbf{31.82} & \textbf{0.8979} & \textbf{28.95} & \textbf{0.8325} & \textbf{30.89} & \textbf{0.9070}\\
\hline
\end{tabularx}
\caption{Comparison against state-of-the-art works on five datasets. The parameter $\sigma$ indicates the noise level of inputs corrupted by additive white Gaussian noise. The best and second best results are in bold and Italic, respectively. Note that for SGNet, the code is not released publicly and their results on Kodak, McMaster and WED-CDM datasets are not reported in their paper.}
\label{table:synthetic}

\end{table*}

\subsection{Corrupted Input as a Weakly Informative Prior}
Data-driven approaches are generally promising when a test image shares similar characteristics with the training set. However, their performances are limited when the input is considerably different, e.g. having a different type of noise. Inspired by the ``noise2noise'' \cite{lehtinen2018noise2noise} and ``mosaic2mosaic'' \cite{ehret2019joint} algorithms, we further improve our trained model by taking the corrupted input as a weakly informative prior, i.e. replacing $y_j$ by $\tilde{x}_j$ during fine-tuning and using a smaller $\lambda$ to indicate the increased uncertainty. However, it comes with an underlying problem: the network may merely learn an identity mapping function, i.e. predicting $\tilde{x}_j$ given $\tilde{x}_j$.\par

It can be observed that in smooth regions, pixels share a strong spatial correlation. Therefore, we tackle the above issue by an alternative scheme, replacing $y_j$ by a random pixel $\tilde{x}_{j+t}$ in a small patch $\{\tilde{x}_{j+t}\}_{t=-\lfloor p^2/2 \rfloor}^{\lfloor p^2/2 \rfloor}$ centered at pixel $\tilde{x}_j$, where $p$ denotes the patch size. It is possible that $\tilde{x}_{j+t}$ differs a lot to $\tilde{x}_j$ in texture-rich regions, requiring another step to exclude the outlying $\tilde{x}_{j+t}$. We achieve this by applying a simple filter: if the value $\tilde{x}_{j+t}$ falls outside the confidence interval $(\tilde{x}_j-2\sigma_j, \tilde{x}_j+2\sigma_j)$, this informative prior $\tilde{x}_{j+t}$ is masked from computing the fine-tuning ELBO loss, i.e.,
\begin{equation}
    \mathcal{L}_{ft} =  \sum_{j=1}^{3N} \mathds{1}(\tilde{x}_{j+t} \in (\tilde{x}_j-2\sigma_j, \tilde{x}_j+2\sigma_j))\mathcal{L}_j(\bm{w};\bm{\tilde{x}}),
\end{equation}
where $\mathds{1}(\cdot)$ denotes an indicator function, and $\mathcal{L}_j(\bm{w};\bm{\tilde{x}})$ can be computed with $j$-indexed components in Equation (\ref{ELBO}), (\ref{eq.Dkl}), (\ref{eq.Expectation}) after replacing $y_j$ by $\tilde{x}_{j+t}$.

\section{Illustrative Experimental Results}
To show the effectiveness of our framework, we conduct extensive experiments with both synthetic datasets and realistic raw data. We focus on the Bayer pattern, which has been the dominating choice among various CFA patterns.

\subsection{Network Architecture}
In previous section, $q_{\bm{w}}(\bm{z},\bm{\sigma}^2|\bm{\tilde{x}})$ represents the network taking $\bm{\tilde{x}}$ as the input. However, the original input is actually $\bm{x}$, and the bilinear interpolation process from $\bm{x}$ to $\bm{\tilde{x}}$ can be considered as part of the job done by the network. Therefore, our network is trained to learn a mapping function $q_{\bm{w}}(\bm{z},\bm{\sigma}^2|\bm{x})$. We use a light-weight network architecture as shown in Figure \ref{fig.network}. The GRDB building module refers to a grouped Residual Dense Block \cite{zhang2018residual}, consisting of dense connected layers and a local feature fusion. A downscaling layer is positioned at the first layer to rearrange a mosaicked input to four quarter-resolution color maps. This rearrangement helps to save memory and speed up the training. Each of the first three convolution layers has 64 filters with $3 \times 3$ kernel size. After that, an upscaling layer is used to unpack the features back to full-resolution. The last convolution layer produces 12 feature maps with $3 \times 3$ kernel size. These 12 feature maps correspond to four parameters $\bm{\hat{y}}, \bm{ \hat{\lambda}}, \bm{ \hat{\alpha}}, \bm{ \hat{\beta}}$, each of which has 3 maps. The network is implemented using PyTorch framework.

\begin{figure*}[t]
    \newcommand \fwidth{0.125}
    \newcommand \firstwidth{0.083}
    \centering
    \begin{subfigure}[b]{\firstwidth \linewidth}
        \centering
        \includegraphics[width=\linewidth]{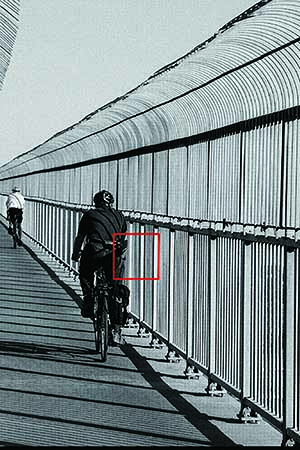}
    \end{subfigure}
    \hfill
    \begin{subfigure}[b]{\fwidth \linewidth}
        \centering
        \includegraphics[width=\linewidth]{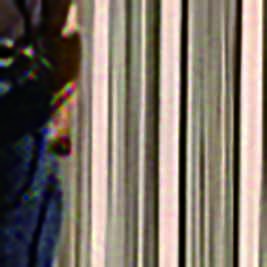}
    \end{subfigure}
    \hfill
    \begin{subfigure}[b]{\fwidth \linewidth}
        \centering
        \includegraphics[width=\linewidth]{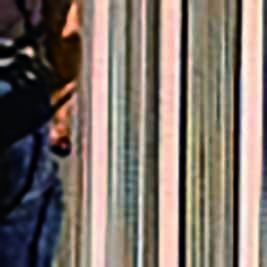}
    \end{subfigure}
    \hfill
    \begin{subfigure}[b]{\fwidth \linewidth}
        \centering
        \includegraphics[width=\linewidth]{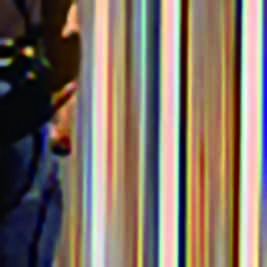}
    \end{subfigure}
    \hfill
    \begin{subfigure}[b]{\fwidth \linewidth}
        \centering
        \includegraphics[width=\linewidth]{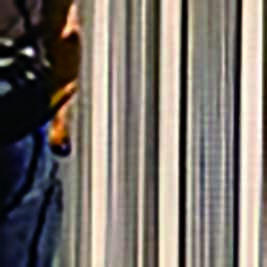}
    \end{subfigure}
    \hfill
    \begin{subfigure}[b]{\fwidth \linewidth}
        \centering
        \includegraphics[width=\linewidth]{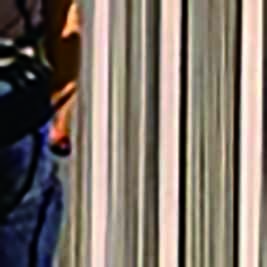}
    \end{subfigure}
    \hfill
    \begin{subfigure}[b]{\fwidth \linewidth}
        \centering
        \includegraphics[width=\linewidth]{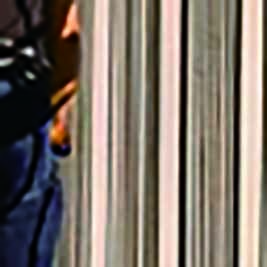}
    \end{subfigure}
    \hfill
    \begin{subfigure}[b]{\fwidth \linewidth}
        \centering
        \includegraphics[width=\linewidth]{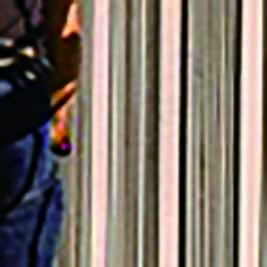}
    \end{subfigure}
    
    \vfill
    \vfill
    
    \begin{subfigure}[b]{\firstwidth \linewidth}
        \centering
        \includegraphics[width=\linewidth]{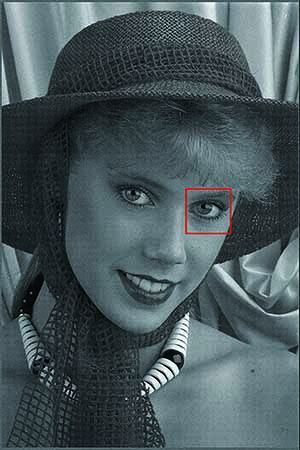}
        \caption{Inputs}
    \end{subfigure}
    \hfill
    \begin{subfigure}[b]{\fwidth \linewidth}
        \centering
        \includegraphics[width=\linewidth]{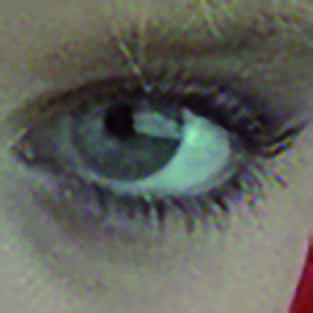}
        \caption{FlexISP}
    \end{subfigure}
    \hfill
    \begin{subfigure}[b]{\fwidth \linewidth}
        \centering
        \includegraphics[width=\linewidth]{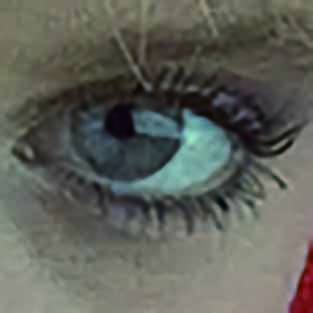}                                   
        \caption{SEM}
    \end{subfigure}
    \hfill
    \begin{subfigure}[b]{\fwidth \linewidth}
        \centering
        \includegraphics[width=\linewidth]{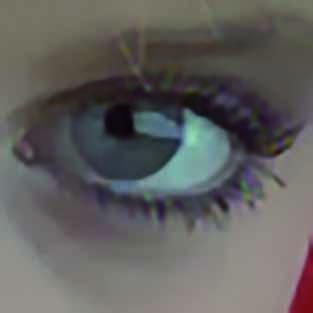}
        \caption{ADMM}
    \end{subfigure}
    \hfill
    \begin{subfigure}[b]{\fwidth \linewidth}
        \centering
        \includegraphics[width=\linewidth]{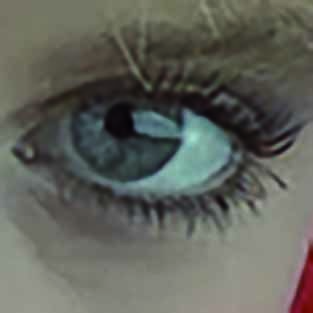}
        \caption{DeepJoint}
    \end{subfigure}
    \hfill
    \begin{subfigure}[b]{\fwidth \linewidth}
        \centering
        \includegraphics[width=\linewidth]{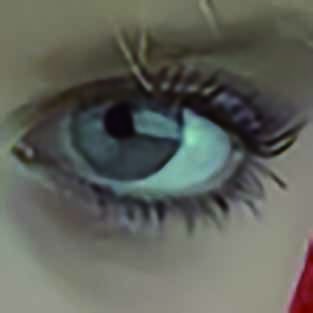}
        \caption{Kokkinos}
    \end{subfigure}
    \hfill
    \begin{subfigure}[b]{\fwidth \linewidth}
        \centering
        \includegraphics[width=\linewidth]{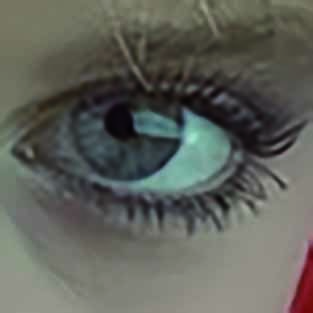}
        \caption{Wild-JDD}
    \end{subfigure}
    \hfill
    \begin{subfigure}[b]{\fwidth \linewidth}
        \centering
        \includegraphics[width=\linewidth]{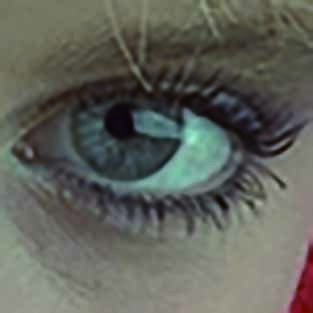}
        \caption{Ground truth}
    \end{subfigure}
    
    \caption{Visual comparison of our method against competing related works. Our reconstructions preserve texture details of high quality without introducing noticeable moire or zipper artifacts.}
    \label{fig.syn_visualiza}
\end{figure*}

\subsection{Experiments on Synthetic Datasets}
We first compare our method with previous works on the synthetic datasets in sRGB space, following the convention without an inverse ISP processing \cite{liu2020joint,klatzer2016learning}. In this experiment, 800 high-resolution images from DIV2K \cite{timofte2017ntire} and 2650 high-resolution images from Flickr2K \cite{lim2017enhanced} are used for training. These images are randomly cropped into $120 \times 120$ patches with a batch size of 128. After the augmentation by flipping and rotation, the noisy mosaicked inputs are generated by applying the Bayer pattern sampling and adding random Gaussian noise in the range of [0, 20]. Unlike most denoising works assuming the i.i.d. noise, which deviates from the practical application, we adopt non-i.i.d Gaussian modeling with spatially variant noise levels, following the work \cite{yue2019variational}. During network training, we empirically set the parameter $\lambda$ as $2e3$ and $\alpha$ as $180.5$ (window size $w$ as 19). The Adam optimizer \cite{kingma2014adam} is used. The learning rate is initialized as $5e\text{-}4$, reduced by a factor of 0.8 when the training meets a plateau in PSNR, with a minimum value of $1e\text{-}4$. The whole training process takes around 5 days on a single RTX 2080Ti GPU.\par 

For testing, five widely-used benchmark datasets are used, including: Kodak\footnote{http://r0k.us/graphics/kodak}, McMaster \cite{zhang2011color}, WED-CDM \cite{tan2017color}, MIT moire \cite{gharbi2016deep} and Urban100 \cite{huang2015single}. These datasets are collected from various devices under different scenarios. Note that the ground truth in these datasets are not perfect either. However, if one method can consistently outperform others across various datasets, its effectiveness can still be validated and approved. For comparison, six existing state-of-the-art works on joint DM\&DN task are adopted, including: FlexISP \cite{heide2014flexisp}, SEM \cite{klatzer2016learning}, ADMM \cite{tan2017joint}, DeepJoint \cite{gharbi2016deep}, Kokkinos \cite{kokkinos2018deep} and SGNet \cite{liu2020joint}. We run their source code for evaluation or directly cite their reported performance if the code is not avaiable. The results are reported in both PSNR and SSIM matrices listed in Table \ref{table:synthetic}.\par 

Overall, our method outperforms all other works quantitatively, though our method is trained for non-i.i.d. noise cases. For both DeepJoint and SGNet, they assume an accurate noise map as known input, which is not reasonable in practice. In contrast, our method is able to perform a truly-blind reconstruction without such a known noise map. To further improve the performance, we adopt a self-ensemble strategy by applying the flipping and rotation on the input to generate 8 augmented inputs. After being processed by the network, 8 outputs are obtained and transformed back to the original geometry, followed by an averaging to get a unified final output. Note that the augmentation on input would break its Bayer pattern, e.g. from RGGB to GRBG. Therefore, a Bayer Preserving Unification is utilized by padding and cropping the image borders \cite{liu2019learning}. We denote our method with self-ensemble as Wild-JDD*.  Qualitative comparison is also provided in Figure \ref{fig.syn_visualiza}. Our reconstructions remove the noise and preserve details pretty well without introducing noticeable artifacts, while other works tend to produce color moire in high-frequency regions. Although Kokkinos \cite{kokkinos2018deep} also has good immunity to those artifacts, the produced images are over-smoothed due to their iterative processing properties. 

\begin{table}
\centering
\begin{tabularx}{0.98\linewidth}{@{\extracolsep{\fill}}lcccc}
\hline
\multirow{2}{*}{Method} & \multicolumn{2}{c}{linear} & \multicolumn{2}{c}{sRGB} \\
                        & PSNR         & SSIM        & PSNR        & SSIM       \\
\hline
JMCDM                   & 37.44        & 0.971       & 31.35       & 0.942      \\
RTF                     & 37.77        & 0.976       & 31.77       & 0.951      \\
FlexISP                 & 38.28        & 0.974       & 31.76       & 0.941      \\
SEM                     & 38.93        &\textit{0.980}       & 32.93       & \textbf{0.960}      \\
DeepJoint               & 38.61        & 0.963       & 32.58       & 0.913      \\
Kokkinos                & 39.29        & 0.975       & 33.37       & 0.930      \\
MMNet20                 & 40.07        & 0.979       & 34.24       & 0.942      \\
DMCNN-VD                & 38.33        & 0.968       & 32.00       & 0.920      \\
DMCNN-VD-Tr             & 40.07        &\textbf{0.981}       & 34.08       &\textit{0.957}      \\
Wild-JDD                    & \textit{40.16}        &\textit{0.980}       & \textit{34.34}       & 0.945      \\
Wild-JDD*                   & \textbf{40.36}        & \textbf{0.981}       & \textbf{34.59}       & 0.947      \\ 
\hline
\end{tabularx}%
\caption{Evaluation on realistic raw data. Our network is trained once using linear data and evaluated on both linear and sRGB space.}
\label{table:msr}
\end{table}

\subsection{Experiments on Realistic Raw Data}
In the previous experiment, we trained and evaluated on sRGB data to enable more comparison with other works. However, \citeauthor{khashabi2014joint} suggested that the evaluation should also be conducted on the raw data and thus proposed a realistic MSR 16-bits benchmark dataset. We retrain our network on their Linear Bayer Panasonic set with 200 images, in the same parameter setting to previous experiment. Table \ref{table:msr} reports our overall better performance in both linear and sRGB space compared to other representative works, including JMCDM \cite{chang2015color}, RTF \cite{khashabi2014joint}, FlexISP \cite{heide2014flexisp}, SEM \cite{klatzer2016learning}, DeepJoint \cite{gharbi2016deep}, Kokkinos \cite{kokkinos2018deep}, MMNet20 \cite{kokkinos2019iterative}, 
DMCNN-VD and DMCNN-VD-Tr \cite{syu2018learning}.

\subsection{Fine-tuning Out-of-distribution Input}
\begin{figure}
\centering
\includegraphics[width=0.78\linewidth]{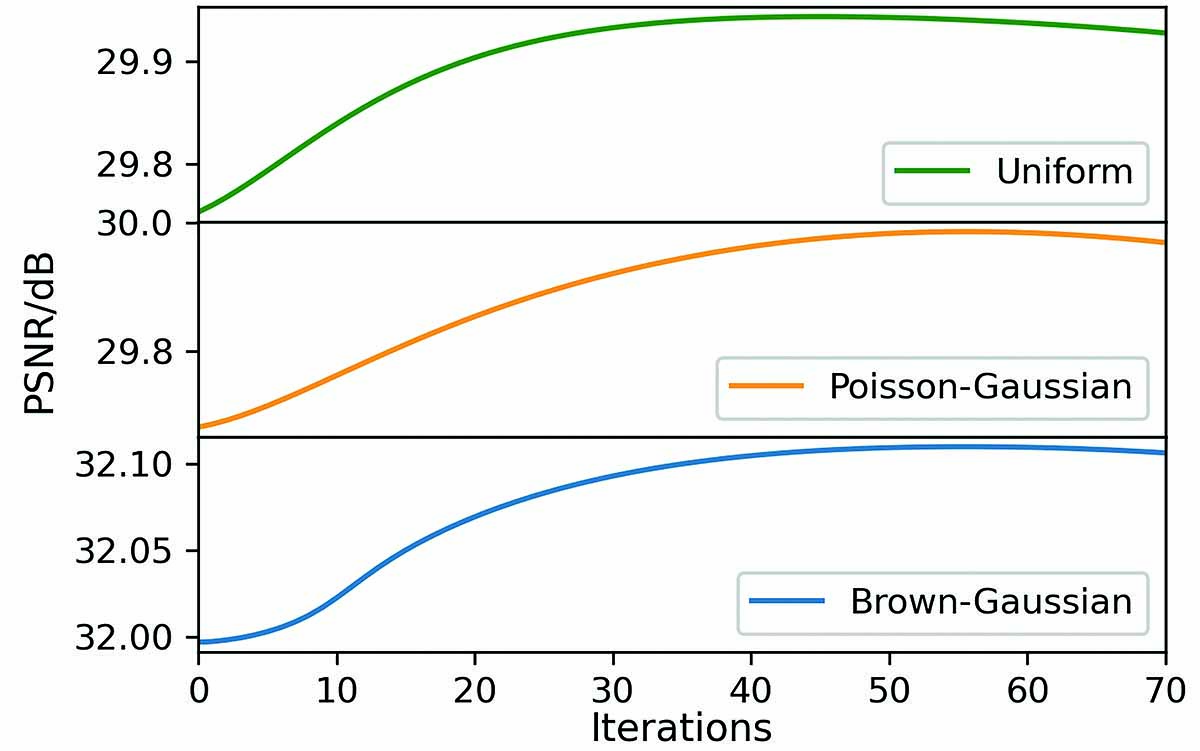}
\caption{Increasing PSNR values when fine-tuning for different corrupted inputs. For each iteration, the updated PSNR values are obtained by averaging them across the McM dataset.}
\label{fig.finetuning}
\end{figure}

 To examine the effectiveness of our fine-tuning strategy, three types of noise are considered, including Uniform, Poisson-Gaussian and Brown-Gaussian. Implementation details are similar to previous experiments except that parameter $\lambda$ is set to be a smaller value $1$, the learning rate decreased to $2e\text{-}6$ and $p$ is empirically set as 3. The results in Figure \ref{fig.finetuning} show a 0.1$\sim$0.3 dB PSNR improvement as the number of iterations increases until roughly 50. Notably, with too many iterations, the performance would drop from its peak. This concern could be eased with the help of no-reference image quality assessment tools \cite{xu2017no}. \par

\begin{table}
\centering
\resizebox{\linewidth}{!}{%
\begin{tabular}{lcccccc}
\hline
\multirow{2}{*}{Method}       & \multicolumn{2}{c}{Kodak}                  & \multicolumn{2}{c}{McMaster}               & \multicolumn{2}{c}{WED-CDM}                \\
& \multicolumn{1}{l}{PSNR} & SSIM            & \multicolumn{1}{l}{PSNR} & SSIM            & \multicolumn{1}{l}{PSNR} & SSIM            \\ \hline
MSE & 33.77 & 0.9124 & 33.45 & 0.9096 & 33.37 & 0.9233 \\
ELBO  & \textbf{33.81}           & \textbf{0.9127} & \textbf{33.53}           & \textbf{0.9123} & \textbf{33.44}           & \textbf{0.9244} \\ \hline
\end{tabular}%
}
\caption{Comparison of ELBO against the MSE on synthetic datasets with noise level $\sigma=10$.}
\label{tab:elbo_mse}
\end{table}

\subsection{Ablation Study}

\subsubsection{ELBO versus MSE} When setting the parameter $\lambda$ to be large enough, our ELBO loss would degrade to a commonly used MSE loss, which assumes the dataset ground truth to be a perfect target. The superiority of using ELBO against MSE loss is validated in Table \ref{tab:elbo_mse}, where a consistent PSNR improvement can be observed. This slight improvement comes from capturing the mild uncertainty embedded in the collected ground truth during training. \par

\begin{figure}
\centering
\includegraphics[width=0.85\linewidth]{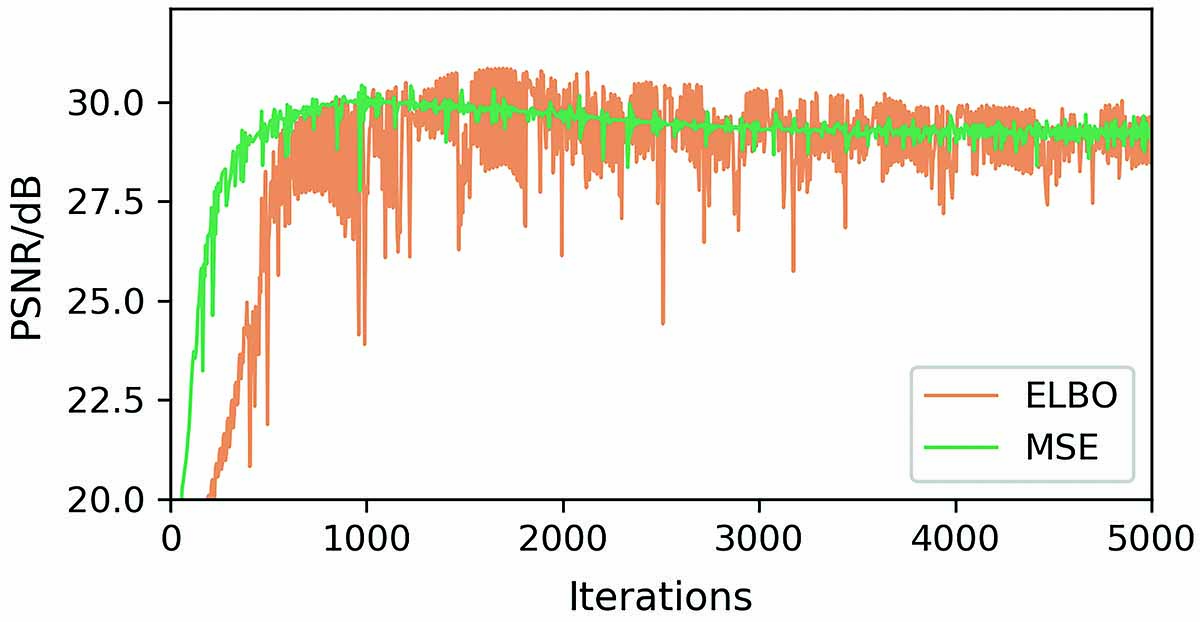}
\caption{Comparison of ELBO against MSE on learning a cartoon image.}
\label{fig.overfitting}
\end{figure}

We also conduct an additional experiment to compare their optimization process. We take a single image from the Cartoon Set \cite{royer2020xgan} as $\bm{z}$. Then a Bayer pattern mosaicking and AWGN noise are applied to this image, followed by a bilinear interpolation to obtain a corrupted version $\bm{\tilde{x}}$. This corrupted image $\bm{\tilde{x}}$ is set to be the learning objective of a neural work. As described in the work Deep Image Prior \cite{ulyanov2018deep}, a neural network tends to learn the clean signal faster than learning the random noise. Therefore, we can see that in Figure \ref{fig.overfitting}, the PSNR curve of MSE increases and then decreases. When ELBO comes into play, its PSNR curve fluctuates roughly around the curve of MSE. This fluctuation results from the interaction between ELBO's restoration term and ELBO's regularization terms. This is consistent to our expectation, that the neural network is aware of the uncertainty affiliated with the target $\bm{\tilde{x}}$ instead of treating $\bm{\tilde{x}}$ as the absolute learning target. Therefore, training with our ELBO loss can achieve a higher intermediate peak of the PSNR curve. \par

\subsubsection{Effect of the mask} 
\begin{figure}
    \centering
    \begin{subfigure}[b]{0.25\linewidth}
        \includegraphics[width=\linewidth]{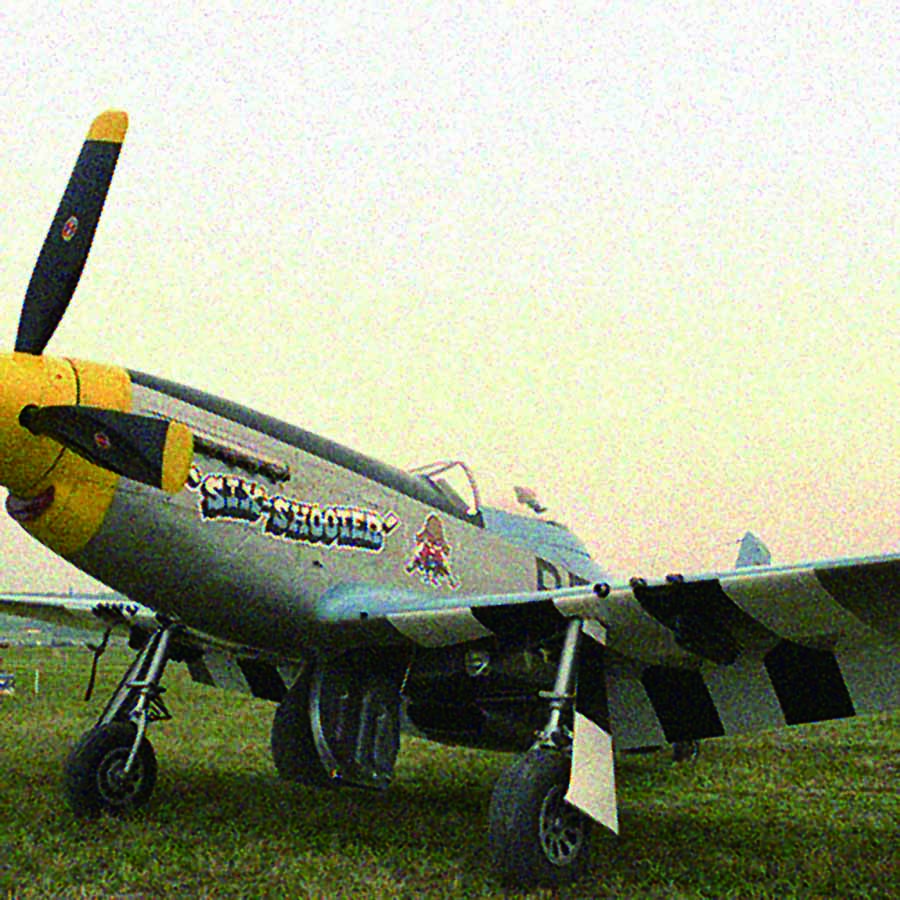}
        \caption{Input $\bm{\tilde{x}}$}
    \end{subfigure}\hfill
    \begin{subfigure}[b]{0.25\linewidth}
        \includegraphics[width=\linewidth]{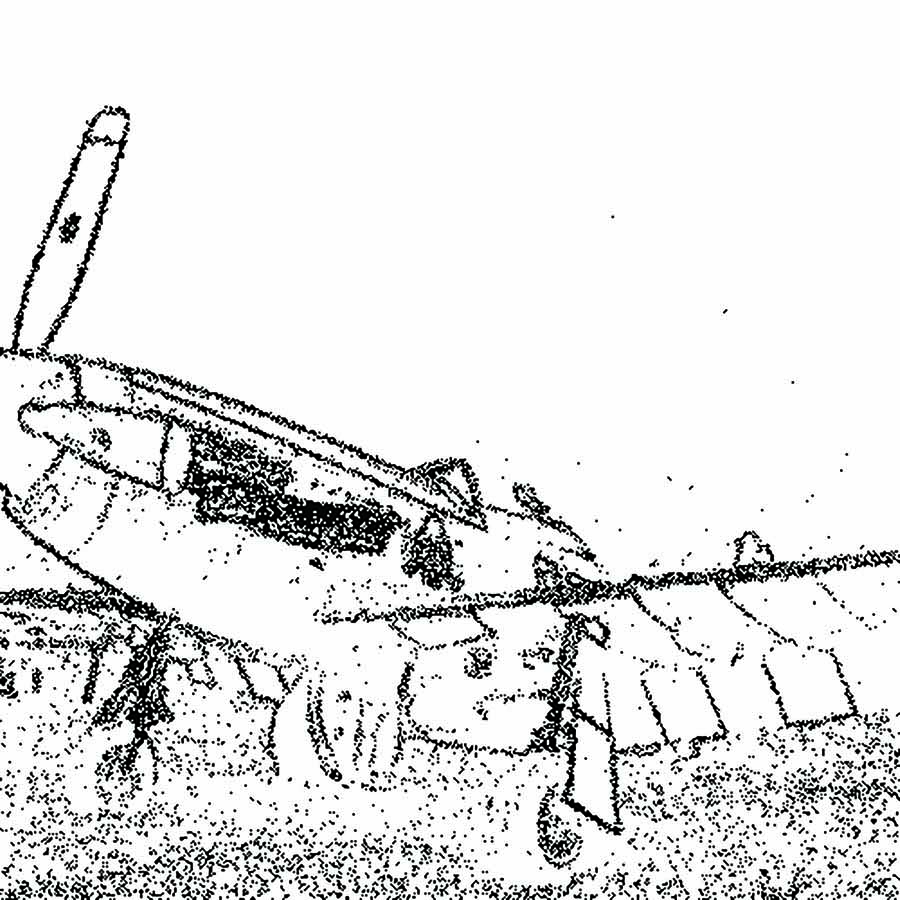}
        \caption{Mask}
        \label{subfig.mask}
    \end{subfigure}\hfill
    \begin{subfigure}[b]{0.4\linewidth}
        \begin{subfigure}[b]{0.3\linewidth}
        \begin{overpic}[width=\textwidth]{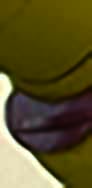}
        \put(15,105) {\scriptsize GT}
        \end{overpic}
        \end{subfigure}\hfill
        \begin{subfigure}[b]{0.3\linewidth}
        \begin{overpic}[width=\textwidth]{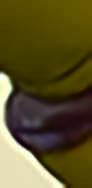}
        \put(15,105) {\scriptsize FT}
        \end{overpic}
        \end{subfigure}\hfill
        \begin{subfigure}[b]{0.3\linewidth}
        \begin{overpic}[width=\textwidth]{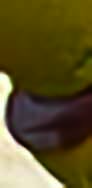}
        \put(-15,105) {\scriptsize FT w/o mask}
        \end{overpic}
        \end{subfigure}
        \caption{Visual comparison}
        \label{subfig.visual_c}
    \end{subfigure}

    \caption{Fine-tuning with and without the mask.}
    \label{fig.ft_mask}
\end{figure}

Using the mask during fine-tuning can effectively avoid edges getting blurred. As shown in Figure \ref{fig.ft_mask} (\subref{subfig.mask}), a mask map generated by our simple confidence interval scheme is able to outline those edges. In Figure \ref{fig.ft_mask} (\subref{subfig.visual_c}), fine-tuning with such a mask preserves the sharp edges more faithfully than fine-tuning without the mask.

\section{Conclusion}
We have presented the Wild-JDD, a novel learning framework for joint demosaicking and denoising tasks. We identify the ground truth uncertainty issues, formulate a two-stage data degradation process and derive an ELBO loss for optimization. We also propose a simple but effective fine-tuning strategy for out-of-distribution input. Comprehensive experiments demonstrate the effectiveness of our method. Wild-JDD not only outperforms state-of-the-art solutions in terms of both statistical and perceptual quality by a clear margin, but also provides good interpretability, where the restoration term and the rest regularization terms coexist to account for the learning target uncertainty. We hope that Wild-JDD will inspire more future research to study the effective training under the ground truth uncertainty in image-to-image translation tasks. \par

\section{Acknowledgement}
This work was supported, in part, by Hong Kong General Research Fund under grant number 16200120.

\bibstyle{aaai21}
\bibliography{ref}

\newpage
\input{supplementary}

\end{document}

%% file: supplementary.tex
\section{Supplementary Materials}
We provide the derivation details of the evidence lower bound (ELBO) loss, comprising a KL-divergence term and an expectation term. For convenience, denote $q_{\bm{w}}(\bm{z},\bm{\sigma}^2|\bm{\tilde{x}})$ as $q(\bm{z},\bm{\sigma}^2)$.
\subsection{Derivations of the KL-divergence term}

\begin{equation*}
     \begin{aligned}
    &\quad D_{KL}(q(\bm{z},\bm{\sigma}^2)||p(\bm{z},\bm{\sigma}^2)) \\
    &=\sum_{j=1}^{3N}\left\{
    D_{KL}(q(z_j, \sigma_j^2)||p(z_j,\sigma_j^2))
    \right\}\\
    &=\sum_{j=1}^{3N}\left\{
    \int_{\mathbb{R}_+} \int_{\mathbb{R}} q(z_j|\sigma_j^2)q(\sigma_j^2) 
    \log{\frac{q(z_j|\sigma_j^2)q(\sigma_j^2) }{p(z_j|\sigma_j^2)p(\sigma_j^2) }}
    dz_jd\sigma_j^2
    \right\}\\
    &=\sum_{j=1}^{3N}\left\{
    \int_{\mathbb{R}_+} q(\sigma_j^2) \int_{\mathbb{R}} q(z_j|\sigma_j^2)
    \log{\frac{q(z_j|\sigma_j^2)}{p(z_j|\sigma_j^2)}} 
    dz_jd\sigma_j^2
    \right.\\
    &\qquad\qquad\left.
    +\int_{\mathbb{R}_+} q(\sigma_j^2) \log{\frac{q(\sigma_j^2) }{p(\sigma_j^2) }} \int_{\mathbb{R}} q(z_j|\sigma_j^2)
    dz_jd\sigma_j^2
    \right\}\\
    &=\sum_{j=1}^{3N}\left\{
    \mathbb{E}_{\sigma_j^2 \sim \Gamma^{-1}(\hat\alpha, \hat\beta_j)}\left[   
    D_{KL}(q(z_j|\sigma_j^2)||p(z_j|\sigma_j^2))
    \right]
    \right.\\
    &\qquad\qquad\qquad\qquad\qquad\qquad\ \ \left.
    +D_{KL}(q(\sigma_j^2)||p(\sigma_j^2))
    \right\}.
    \end{aligned}
\end{equation*}
Using the KL-divergence between two Gaussian distributions, we have

\begin{equation*}
    \begin{aligned}
    &D_{KL}(q(z_j|\sigma_j^2)||p(z_j|\sigma_j^2)) \\
    =&\frac{\lambda}{2\sigma_j^2}(y_j-\hat{y}_j)^2+\frac{\lambda}{2\hat{\lambda}_j}-\frac{1}{2}\log\frac{\lambda}{\hat{\lambda}_j}-\frac{1}{2},
    \end{aligned}
\end{equation*}
hence,
\begin{equation*}
    \begin{aligned}
    &\mathbb{E}_{\sigma_j^2 \sim \Gamma^{-1}(\hat\alpha, \hat\beta_j)}\left[
    D_{KL}(q(z_j|\sigma_j^2)||p(z_j|\sigma_j^2))
    \right]\\
    =&\frac{\lambda\hat{\alpha}_j}{2\hat{\beta}_j}(y_j-\hat{y}_j)^2+\frac{\lambda}{2\hat{\lambda}_j}-\frac{1}{2}\log\frac{\lambda}{\hat{\lambda}_j}-\frac{1}{2}.
    \end{aligned}
\end{equation*}
And using the KL-divergence between two Inverse-Gamma distributions, we have
\begin{equation*}
    \begin{aligned}
    &D_{KL}(q(\sigma_j^2)||p(\sigma_j^2))\\
    =&\alpha_j\log{\frac{\hat{\beta}_j}{\beta_j}}  +\log{\frac{\Gamma(\alpha_j)}{\Gamma(\hat{\alpha}_j)}}+(\hat{\alpha}_j-\alpha_j)\psi(\hat{\alpha}_j)-(\hat{\beta}_j-\beta_j)\frac{\hat{\alpha}_j}{\hat{\beta}_j},
    \end{aligned}
\end{equation*}
where $\Gamma(\cdot), \psi(\cdot)$ denote the Gamma and Di-gamma function respectively. Therefore, we have
\begin{equation*}
    \begin{aligned}
    &\quad D_{KL}(q(\bm{z},\bm{\sigma}^2)||p(\bm{z},\bm{\sigma}^2)) \\
    &= \sum_{j=1}^{3N}\left\{
    \frac{\lambda\hat{\alpha}_j}{2\hat{\beta}_j}(y_j-\hat{y}_j)^2+\frac{\lambda}{2\hat{\lambda}_j}-\frac{1}{2}\log\frac{\lambda}{\hat{\lambda}_j}+\alpha_j\log{\frac{\hat{\beta}_j}{\beta_j}}\right.\\
    &\quad  \left.-\frac{1}{2}+\log{\frac{\Gamma(\alpha_j)}{\Gamma(\hat{\alpha}_j)}}+(\hat{\alpha}_j-\alpha_j)\psi(\hat{\alpha}_j)-(\hat{\beta}_j-\beta_j)\frac{\hat{\alpha}_j}{\hat{\beta}_j} \right\}.
    \end{aligned}
\end{equation*}


\subsection{Derivations of the expectation term}

\begin{equation*}
     \begin{aligned}
    &\quad \mathbb{E}_{q(\bm{z},\bm{\sigma}^2)}\left[\log p(\bm{\tilde{x}} | \bm{z},\bm{\sigma}^2)\right] \\
    &=\sum_{j=1}^{3N} \mathbb{E}_{q(z_j,\sigma_j^2)}\left[\log p(\tilde{x}_j | z_j,\sigma_j^2)\right] \\
    &=\sum_{j=1}^{3N} \mathbb{E}_{q(z_j,\sigma_j^2)}\left[
    -\frac{\log{2\pi}}{2}-\frac{\log{\sigma_j^2}}{2}-\frac{(\tilde{x}_j-z_j)^2}{2\sigma_j^2}
    \right] \\
    &=\sum_{j=1}^{3N}\left\{ 
    -\frac{\log{2\pi}}{2}
    -\mathbb{E}_{q}\left[
    \frac{\log{\sigma_j^2}}{2}\right] 
    -\mathbb{E}_{q}\left[
    \frac{(\tilde{x}_j-z_j)^2}{2\sigma_j^2}\right]
    \right\}\\
    &=\sum_{j=1}^{3N}\left\{
    -\frac{\log{2\pi}}{2}
    -\frac{\log{\hat\beta_j}-\psi(\hat\alpha_j)}{2} \right.\\
    &\quad\  \left. 
    -\mathbb{E}_{\sigma_j^2 \sim \Gamma^{-1}(\hat\alpha, \hat\beta_j)}\left[
    \mathbb{E}_{z_j \sim \mathcal{N}(\hat{y}_j, \sigma_j^{2}/\hat\lambda_j)}\left[
    \frac{(\tilde{x}_j-z_j)^2}{2\sigma_j^2}
    \right]
    \right]
    \right\}\\
    &=\sum_{j=1}^{3N}\left\{
    -\frac{\log{2\pi}}{2}
    -\frac{\log{\hat\beta_j}-\psi(\hat\alpha_j)}{2} \right.\\
    &\qquad\qquad\qquad \  \left. 
    -\mathbb{E}_{\sigma_j^2 \sim \Gamma^{-1}(\hat\alpha, \hat\beta_j)}\left[
    \frac{\sigma_j^2}{2\hat{\lambda}_j^2}+
    \frac{(\tilde{x}_j-\hat{y}_j)^2}{2\sigma_j^2}
    \right]
    \right\}\\
    &=\sum_{j=1}^{3N}\left\{
    -\frac{\log{2\pi}}{2}
    -\frac{\log{\hat\beta_j}-\psi(\hat\alpha_j)}{2} 
     \right.\\
    &\qquad\qquad\qquad\qquad\ \ \  \left.
    -\frac{\hat\beta_j}{2\hat{\lambda}_j^2(\hat\alpha_j-1)}
    -\frac{\hat\alpha_j(\tilde{x}_j-\hat{y}_j)^2}{2\hat\beta_j}
    \right\}.
    \end{aligned} 
\end{equation*}